\documentclass[letterpaper, 10 pt, conference]{ieeeconf}  

\IEEEoverridecommandlockouts                              

\usepackage{cite}
\usepackage{amsmath,amssymb,amsfonts}
\usepackage{algorithmic}
\usepackage{graphicx}
\usepackage{textcomp}
\usepackage{xcolor}
\usepackage{multirow}
\usepackage{makecell}
\usepackage{tabularx}
\usepackage[colorlinks,linkcolor=blue,anchorcolor=blue,citecolor=blue]{hyperref}
\usepackage{amsfonts}
\usepackage{flushend}
\usepackage{verbatim}
\usepackage{array}
\usepackage{soul} 
\usepackage{color, xcolor} 
\usepackage{bbding,pifont,utfsym,fontawesome}
\newcolumntype{C}[1]{>{\centering}p{#1}}
\begin{document}

\title{PoseFusion: Robust Object-in-Hand Pose Estimation with SelectLSTM
\author{Yuyang Tu$^{{\dag1}}$,  Junnan Jiang$^{{\dag2}}$, Shuang Li$^{1}$, Norman Hendrich$^{1}$, Miao Li$^{{\ast2}}$ and Jianwei Zhang$^{{\ast1}}$} 
\thanks{$^1$ Department of Informatics, Universiät Hamburg, Hamburg 20146, Germany. }
\thanks{$^2$ Institute of Technological Sciences, Wuhan University, Hubei, China}
\thanks{$\dag$  The first two authors contributed equally to this work}
\thanks{$\ast$ Corresponding author (e-mail: miao.li@whu.edu.cn, zhang@informatik.uni-hamburg.de)}
\thanks{This research was funded by the German Research Foundation (DFG) and the National Science Foundation of China (NSFC) in project Crossmodal Learning, DFG TRR-169/NSFC.}

}
\maketitle

\begin{abstract}
Accurate estimation of the relative pose between an object and a robot hand is critical for many manipulation tasks. However, most of the existing object-in-hand pose datasets use two-finger grippers and also assume that the object remains fixed in the hand without any relative movements, which is not representative of real-world scenarios. To address this issue, a 6D object-in-hand pose dataset is proposed using a teleoperation method with an anthropomorphic Shadow Dexterous hand. Our dataset comprises RGB-D images, proprioception and tactile data, covering diverse grasping poses, finger contact states, and object occlusions. To overcome the significant hand occlusion and limited tactile sensor contact in real-world scenarios, we propose PoseFusion, a hybrid multi-modal fusion approach that integrates the information from visual and tactile perception channels. PoseFusion generates three candidate object poses from three estimators (tactile only, visual only, and visuo-tactile fusion), which are then filtered by a SelectLSTM network to select the optimal pose, avoiding inferior fusion poses resulting from modality collapse. Extensive experiments demonstrate the robustness and advantages of our framework. All data and codes are available on the project website: \href{https://elevenjiang1.github.io/ObjectInHand-Dataset/}{https://elevenjiang1.github.io/ObjectInHand-Dataset/}.

\end{abstract}

\section{Introduction}

Accurate object-in-hand pose estimation is vital in many applications, such as pick and place, tool usage, and peg-in-hole assembly. In many situations, vision information can be used to obtain the pose of the object precisely. However, vision is strongly influenced by object occlusion and lighting conditions, as well as intrinsic noise from the camera itself. As a counterpart, tactile information can be used to obtain 
contact positions and contact forces, which can be used to refine the pose estimation from the methods that use vision alone. However, such tactile information is usually discrete in space and time-varying if relative movement occurs between the object and the hand. As a result, how to best leverage the power of vision and tactility to obtain an accurate pose estimation of the in-hand object remains an open research question.

Previous work has studied this problem by matching object point clouds and contact points using computational methods \cite{bimbo2012object} and learning-based methods\cite{dikhale2022visuotactile}. Other work investigated the active tactile pose estimation for grasping from a dense clutter \cite{murali2022active}. Visual and optical tactile sensing has also been used to estimate the pose of a fixed object \cite{chaudhury2022using}. Though diverse learning-based methods have been proposed to fuse different sensors to predict 6D poses, they still ignore some fatal problems. First, most existing datasets use two-finger grippers integrated with tactile sensors \cite{murali2022active,bauza2022tac2pose}. The objects are either fixed on the table or held in the robot's hand with the assumption that there is no relative movement between the hand and the objects, and no multi-finger dataset is available. Moreover, for the fusion of the visuo-tactile data, the confidence of the different modalities can vary over time. Directly fusing multi-modal data without considering the effect of single-modal corruption often leads to network collapse.

\begin{figure}[!t]
\centerline{\includegraphics[width=3.5in]{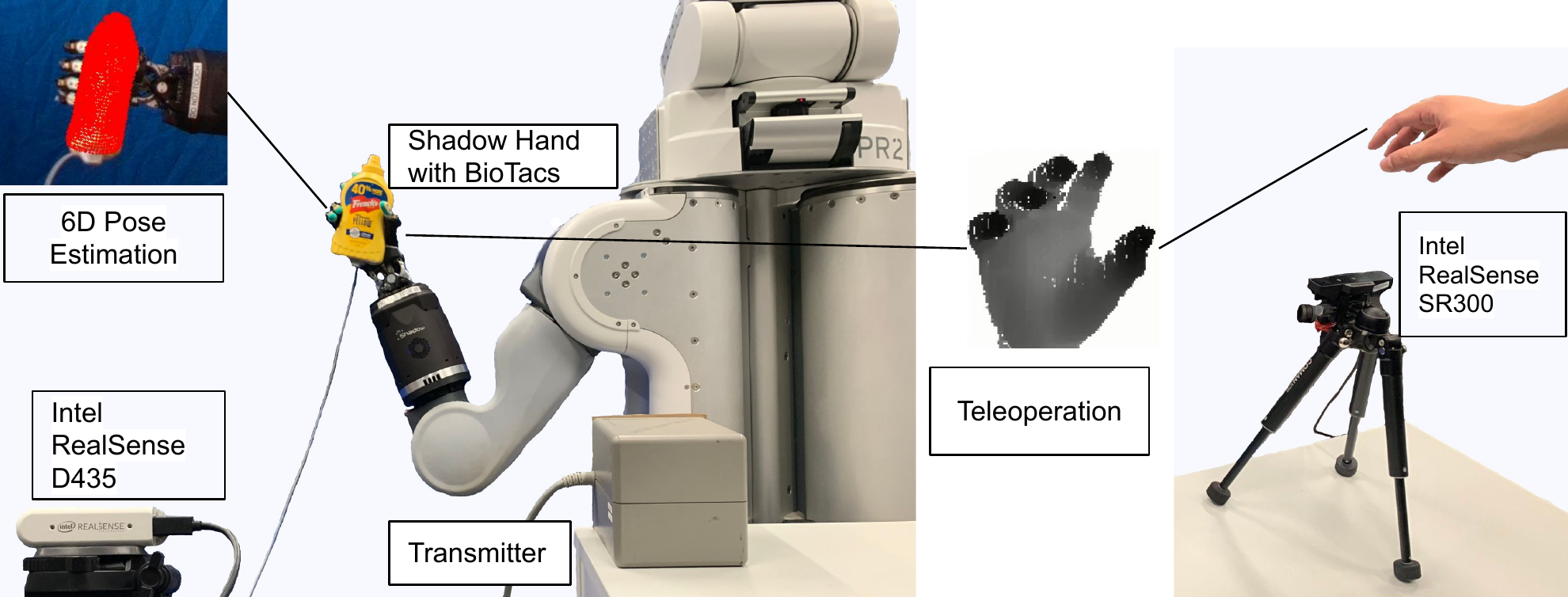}}
\caption{Experimental setup for the ObjectInHand dataset collection. We use an Intel RealSense SR300 camera to record the operator's hand to teleoperate a PR2 robot with Shadow Dexterous hand during grasping and in-hand manipulation tasks (right). The actual dataset (left) includes RGB-D data from a second Intel RealSense D435 camera that observes the Shadow hand and the object. The real-time 6D object poses ground-truth data are obtained from a magnetic tracking system (Polhemus Liberty), with the sensor attached to the manipulation object and the reference transmitter in front of the robot.}
\label{fig1}
\end{figure}

To address the challenges mentioned above, in this work, firstly, we employ a vision-based teleoperation method to manipulate and reorient the object in hand, which allows us to record a novel dataset that includes diverse grasping poses, finger contact states, and object occlusions. Secondly, we propose PoseFusion, a hybrid multi-modal fusion approach that integrates information from visual and tactile perception channels to overcome significant hand occlusion and limited contact between tactile sensors and objects in real-world manipulation scenarios. PoseFusion generates three candidate object poses from three estimators: tactile-only, visual-only, and visuo-tactile fusion. The tactile-only estimator uses only the tactile sensor and dexterous hand proprioception data, while the visual-only estimator uses the RGB-D image data. The visuo-tactile fusion estimator combines the information from both modalities. These three estimators generate candidate poses that are then filtered by a SelectLSTM network to select the optimal pose, avoiding inferior fusion poses resulting from modality collapse. To further analyze errors caused by imperfections in individual sensor channels, we quantify the data quality in our ObjectInHand dataset using the occlusion rate and the number of contacted tactile sensors. Finally, we validate our PoseFusion method in experiments, to estimate in-hand object pose with manually added occlusions or with tactile sensors not touching the object. The illustration and the whole setup of our experiments are shown in Fig. \ref{fig1}.

To the best of our knowledge, the novelty of our work includes the following:
\begin{itemize}
\item[$\bullet$] A real-world dataset is collected using a Shadow hand equipped with BioTac tactile sensors. The dataset takes into account different hand occlusion scenarios in 4 grasping styles and with multiple distances of objects from the camera.
\item[$\bullet$] A novel hybrid network (PoseFusion and SelectLSTM) is proposed to select among the three pose candidates from three different modal pose estimators: tactile-only, vision-only and visuo-tactile fusion. 
Experiments demonstrate this can well address the problem of corrupted data, which may easily be encountered in real scenarios.
\end{itemize}

\section{Related Work}

\subsection{6D Pose Dataset}

Early 6D pose datasets were used for the perception of the environment in static scenes, such as LINEMOD \cite{hinterstoisser2012model} and YCB-Video \cite{xiang2018posecnn}. For easier data acquisition, some work uses simulation to generate synthetic data \cite{jalal2019sidod,tyree20226}. Then, to investigate dynamic scenes, for example, object placement or tool use, object pose in human hand datasets are proposed \cite{Doosti_2020_CVPR,chao2021dexycb}. In robotics, although there are many 6D datasets with objects in hand, these datasets often assume that the object is attached to the end effector \cite{wen2020se} or the desktop \cite{bauza2022tac2pose}. 
Furthermore, most existing datasets that estimate object-in-hand poses only consider ideal situations (e.g. stable grasps without slippage) while using non-humanoid robot hands with simple mechanisms and no tactile sensing involved. In contrast, we provide a dataset in the context of dexterous manipulation by teleoperating a dexterous hand with realistic in-hand object motions.

\subsection{In-hand Pose Estimation}

To perform manipulation tasks such as object orientation and tool use \cite{andrychowicz2020learning,chen2022system}, it is necessary to obtain and track the object in-hand pose during task execution. Since the tactile-based approaches \cite{kuppuswamy2019fast,bauza2022tac2pose} degrade if the tactile sensor loses contact with the object, and the vision-based \cite{wen2020robust} approaches degrade due to object occlusions from hand and fingers or clutter objects, it is a common idea to fuse the tactile and vision data to perform in-hand pose estimates. A DeepGate structure was used to determine the importance of vision and tactile modalities in \cite{anzai2020deep}, and a fusion method of semantic tactility and visual point cloud for pose estimation was proposed in \cite{dikhale2022visuotactile}. An active perception method was also used to estimate the pose of unknown objects \cite{kelestemur2022tactile}. This prior work has extensively explored the problem in the ``static grasp'' context, where the robot hand grasps an object firmly without relative movement. In contrast to the case of static grasps, our work tackles the more challenging problem of estimating the in-hand object pose during dexterous manipulation.

\begin{figure*}[!t]
\centerline{\includegraphics[width=\linewidth]{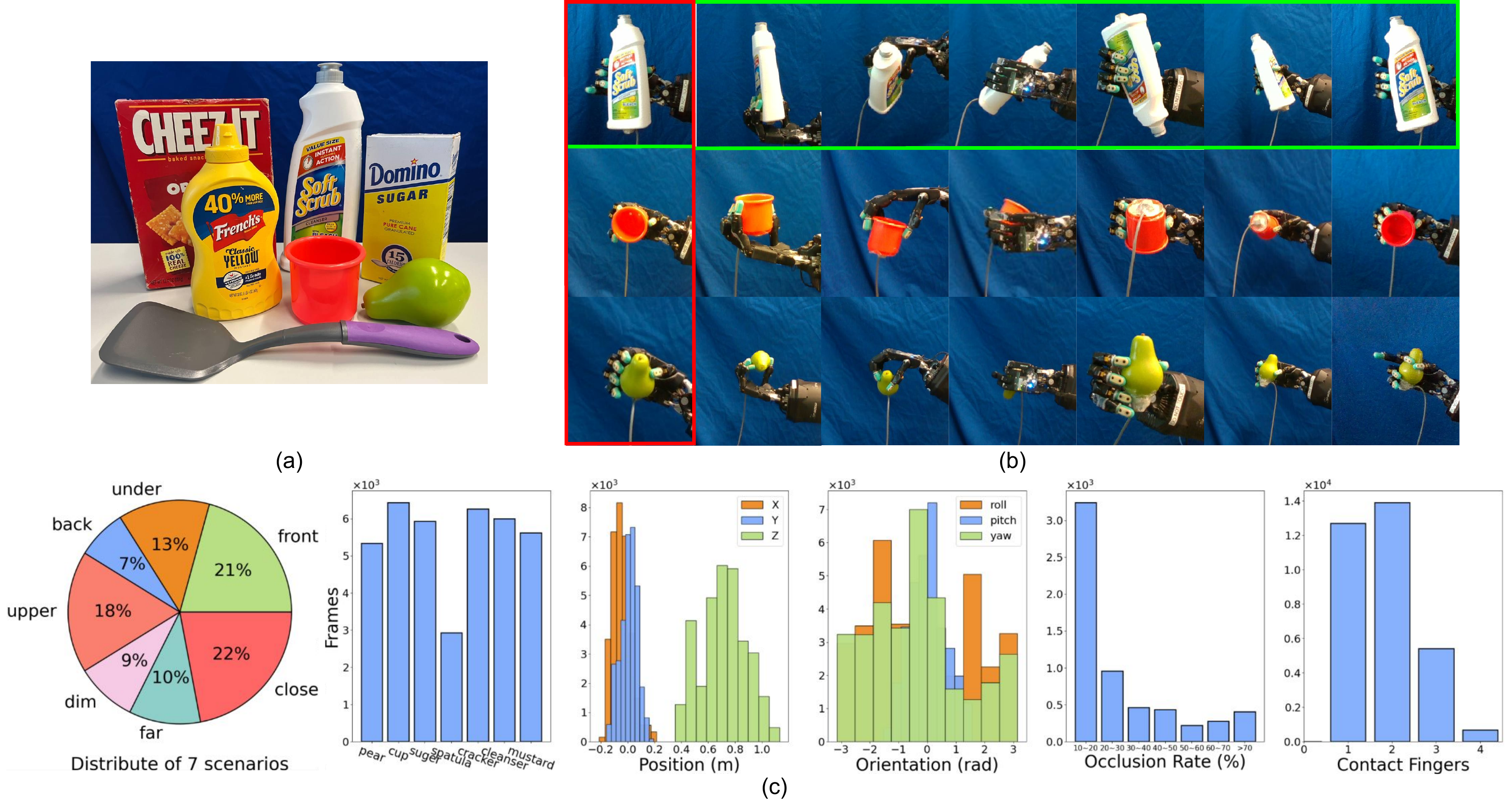}}
\caption{(a) The 7 different objects in the ObjectInHand dataset. (b) Different initial robot hand poses in the ObjectInHand dataset. The first column is the baseline scenario, the object is in front of the hand (front), followed by other derived scenarios, object on the hand (upper), object under the hand (under), object at the back of the hand (back), object close to the camera (close), object far from the camera (far), bad illumination (dim). (c) Data distribution of ObjectInHand Dataset.}
\label{fig2}
\end{figure*}

\subsection{Resolving Single-Modal Corruption}

Recently, multi-modal fusion methods have been making rapid progress in the autonomous driving and robotic community \cite{bai2022transfusion,lee2019making,cui2021deep,wang2021pointaugmenting}. However, multi-modal data also increases the possibility of incorrect data when faced with low-quality input data. In the context of dexterous robotic manipulation with re-grasping and finger jittering, fingertip tactile sensors will only be in contact with the object at certain times, while vision data will frequently be partially or even fully occluded by the dexterous hand.
Therefore, the effect of data fusion may be less effective than focusing on a single modality. To solve such data corruption problems, a gated network to determine the reliability of each modality and assign them to the pose estimator in the next stage was proposed in \cite{anzai2020deep}. 
A highly scalable ``mixture of experts'' approach was presented in \cite{45929}.
In addition, with the development of generative models, some works began to use Generative Adversarial Networks (GANs) to generate a supplement of realistic data for the impacted sensor modality \cite{wu2018multimodal,lee2021detect}. However, all the methods above only yield a single result, in the hope that the network structure in the middle layer can adaptively avoid collapsing feature-level data. In this paper, we select the output results of each single and feature-level fused modality by performing Gate Learning/Selection in the output layer. 

\section{ObjectInHand Dataset}

Our ObjectInHand dataset aims to record the combination of
RGB-D camera images, tactile data, dexterous hand joints data, and target object pose data 
during prototypical in-hand manipulation tasks. 
We use a teleoperation method to perform dexterous manipulation to change the pose of the in-hand object constantly. To ensure the diversity of the dataset, we record the data of 7 objects under 4 different grasping styles and at multiple distances from the camera and with different illuminations.

\subsection{Data Collection Setup}

As shown in Fig. \ref{fig1}, our data collection system consists of both a teleoperation part and a data recording part. 
We utilize a recent vision-based teleoperation method, TeachNet,
to perform the in-hand object manipulations
\cite{DBLP:conf/icra/LiMLGR0S019}.
The human teleoperator moves their fingers in front of the Intel Real\-Sense SR300 depth sensor. Controlled by the deep network, the Shadow hand imitates the human hand and manipulates the in-hand object in real-time. During our experiments, the human teleoperator always tries to generate different grasping poses of the robot hand, so as to yield diverse object poses. 

In the data collection part, we record RGB-D images with a resolution of 640$\times$480 pixels
from an Intel RealSense D435 depth camera. 
Tactile data is provided by Syntouch BioTac sensors \cite{wettels2008biomimetic} 
mounted on each fingertip of the hand.
Each BioTac sensor generates a 19-dim analog signal corresponding to the deformation of the elastic sensor surface 
under contact forces. 
The 6D poses of all Shadow hand links, including the BioTac sensors, are recorded in terms of ROS TF \cite{ros,foote2013tf}. 

We calibrate the camera and the PR2 robot on which the Shadow hand is mounted with a hand-eye calibration method proposed in \cite{tsai1989new}. The post-processing of tactile information, including filtering and finger contact detection is also provided in our dataset. 
To record the ground truth 6D pose of the manipulated object, we attached an electromagnetic 6D pose tracking sensor (Polhemus Liberty) to the bottom of the object. The ground truth pose can be tracked with about 0.76\,mm static accuracy in position and 0.15 degrees accuracy in orientation. Thus we only need to use ICP \cite{121791}, which takes the object point cloud in the camera frame and the ground truth point cloud as input to calibrate the object's ground truth pose once for the first frame. In this manner, we can avoid manually annotating all the video frames to collect the data more efficiently.

\begin{figure}[b]
\centerline{\includegraphics[width=3.5in]{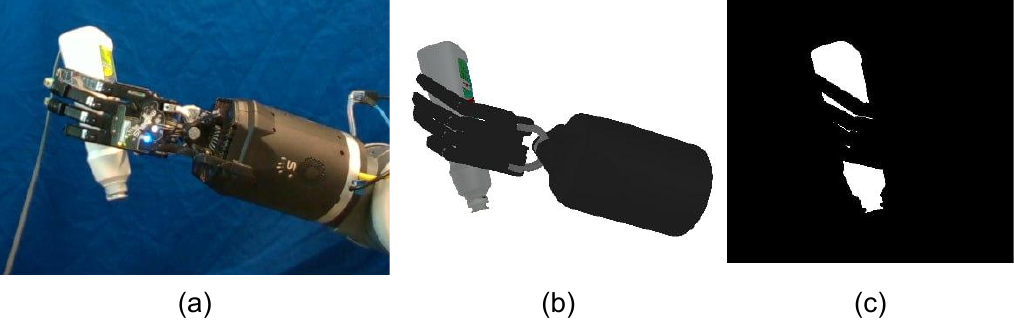}}
\caption{Occlusion calculation setup. (a) Image from ObjectInHand Dataset, (b) Synthetic image with the same configuration in PyBullet, (c) Synthetic camera segmented image of the object.} 
\label{fig3}
\end{figure}

\begin{figure*}[!t]
\centering
\includegraphics[width=\linewidth]{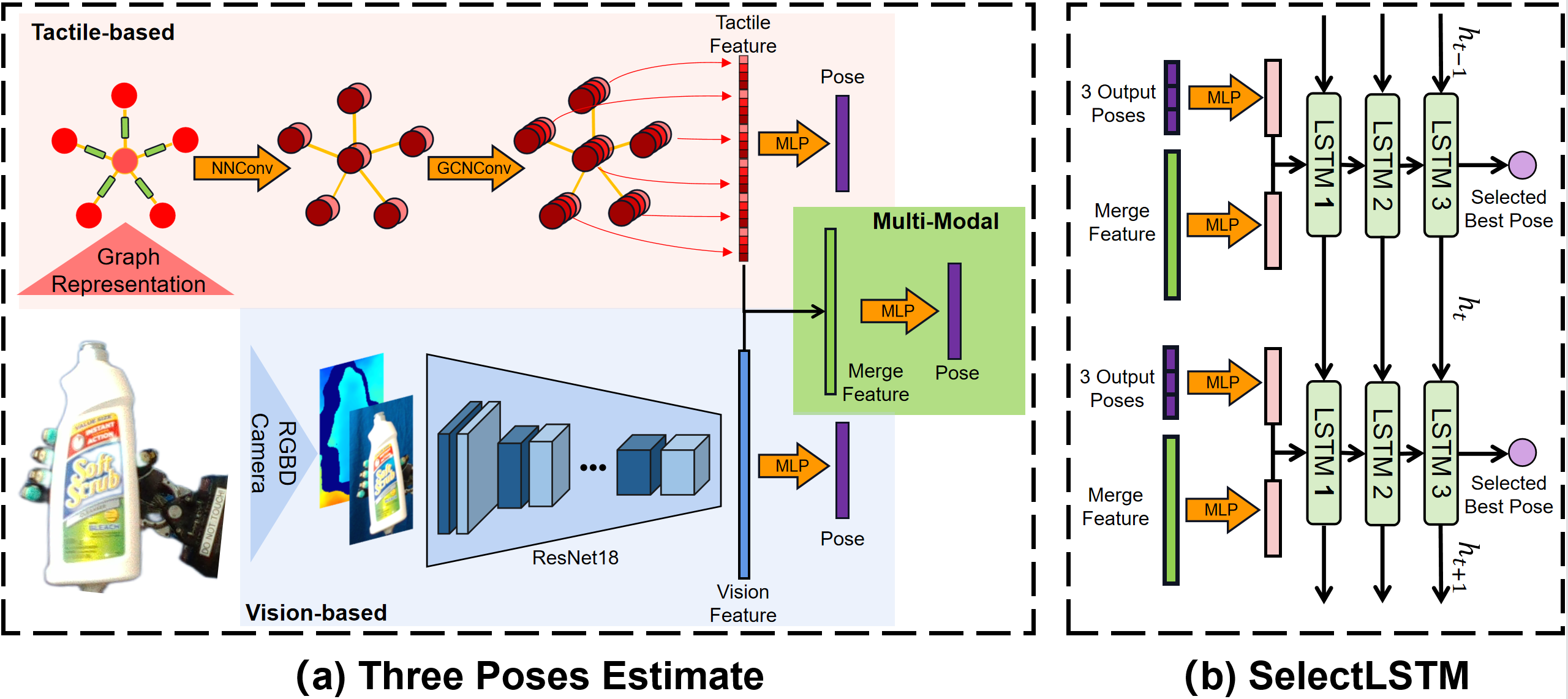}
\caption{The whole pipeline of PoseFusion. First, tactile data including electrode reading and fingertip pose, and vision data including RGB and depth images are processed separately for feature extraction and pose estimation. Then the merged features, including tactile and vision features, are also used for pose estimation. Finally, the three output poses and the merged features are fed into SelectLSTM for the best pose selection among the three poses, which aims to avoid the error of single-modality data collapse.}
\label{fig4}
\end{figure*}

\subsection{Dataset Characteristics}%

We choose 7 objects from the YCB dataset since they are widely accessible \cite{calli2015benchmarking}. The selected objects include a cracker box, sugar box, mustard bottle, pear, bleach cleanser bottle, spatula, and cup, as shown in Fig. \ref{fig2}(a). The spatula is difficult to touch with a fingertip in tactile sensing, while the cup and pear have low-texture surfaces. Additionally, the bleach cleanser bottle, cracker box, sugar box, and mustard bottle vary in size and weight distribution. 

To include diverse dexterous manipulation scenarios into the ObjectInHand dataset, we start to manipulate each object in 4 different grasping poses and correspondingly different initial object positions relative to the robot hand. The objects are positioned in front of the hand, on top of the hand, under the hand, and at the back of the hand, as shown in Fig. \ref{fig2}(b). Also, the dataset includes scenarios based on the object in front of the hand both close to and far from the camera, as well as a scenario where the lab illumination is turned off. In the dataset, we marked the occlusion rate of each data frame. To easily calculate the occlusion rate, we put the models of the YCB object and the robot hand into a PyBullet simulation environment to get the synthetic camera segmentation mask (See Fig. \ref{fig3}). The occlusion rate is defined as the percentage of the object in the image hidden by the robot hand. 

We also marked the estimated contact points of the tactile sensors with the object in each data frame. The point-of-contact estimation method was suggested by\cite{JonetzkoMaster2017}. Notice that we do not use the estimated contact points in our 6D pose estimation method (see below), but the data can be accessed in our dataset. Fig. \ref{fig2}(c) shows the distribution of our dataset, including the distribution of 7 scenarios, XYZ positions, roll-pitch-yaw orientation, occlusion rate, number of contacted tactile fingers, and number of image frames for each object.

\section{PoseFusion}

Our PoseFusion network mainly contains three estimators for candidate pose estimation, along with a SelectLSTM for selecting the best pose. First, a graph-based data structure is used to represent the tactile data. Then, we obtain three candidate poses using three pose estimators: tactile-only, vision-only, and visuo-tactile data. Finally, a subsequent selection network (SelectLSTM) is used to select the best pose among the three poses, aiming to deal with problems such as modal data collapse and sudden changes in the estimated results. The whole pipeline of our PoseFusion is shown in Fig. \ref{fig4}.

\begin{figure}[!b]
\centerline{\includegraphics[width=3.4in]{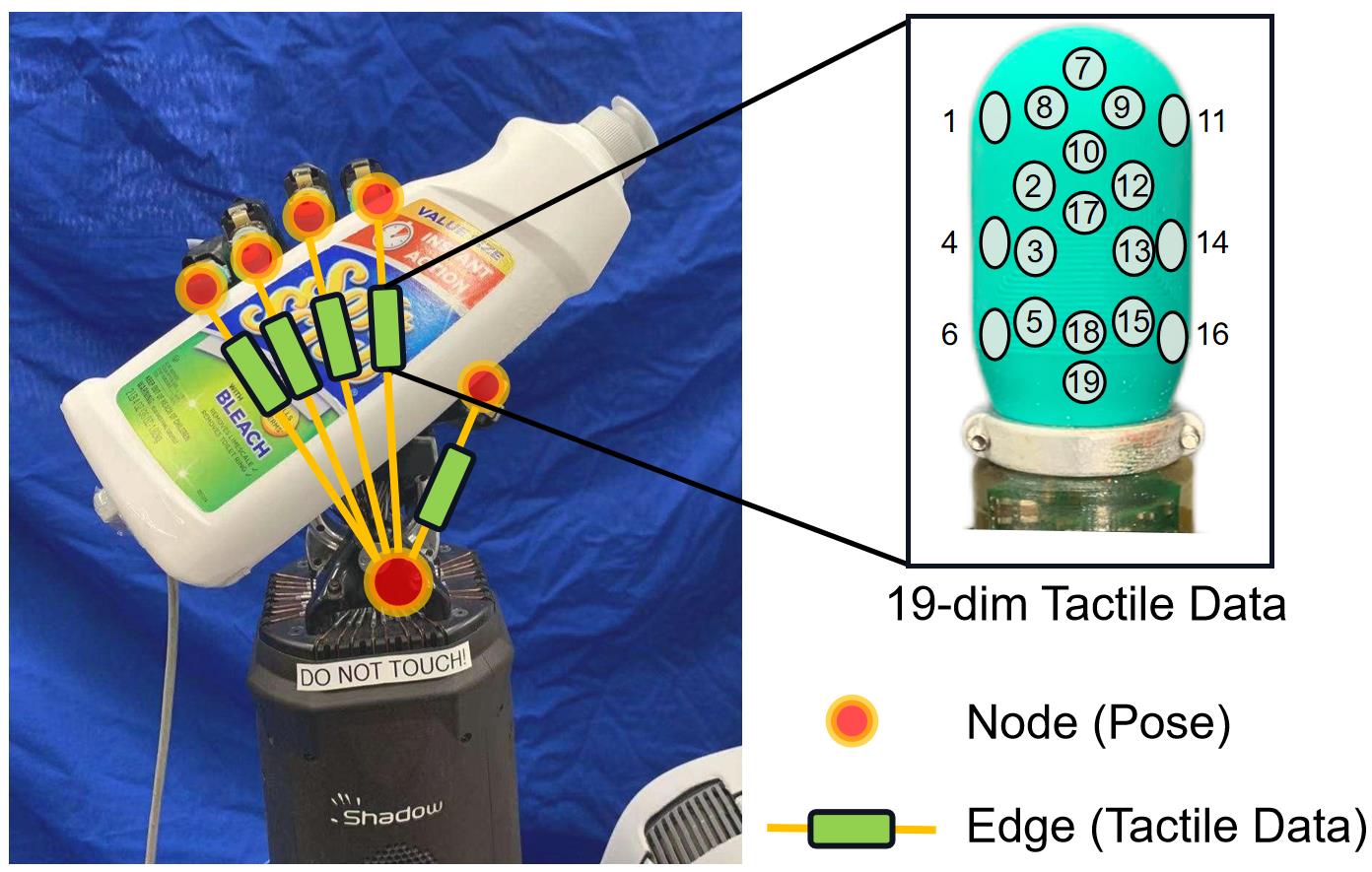}}
\caption{Graph-based representation of dexterous hand. The poses from the 5 fingertips to the hand palm are represented by 5 distal nodes, while the pose from the hand palm to the camera is represented by the central node. And the 19-dim electrode readings from the tactile sensor on 5 fingertips are used as the edge features connecting each fingertip node to a hand palm node.} 
\label{graph-representation}
\end{figure}

\subsection{Graph Representation of the Dexterous Hand}\label{Graph-Representation-subsection}
The data from the dexterous hand contains 7-dim poses (position and quaternion) transformed between each fingertip and the hand palm, and also 19-dim of tactile data from each tactile sensor on the fingertip. To represent and extract features from these data, we use a graph-based data structure, as shown in Fig. \ref{graph-representation}. Specifically, we take the the fingertip pose relative to the palm and the palm pose relative to the camera frame coordinate system as the node information, one transform for one node. The 19 electrode readings from the tactile sensor are used as edge features connecting the corresponding finger nodes to the palm nodes. Finally, an undirected graph with 6 nodes and 5 edges is formed, which can be used for feature extraction by graph convolution, as described in detail in \ref{6D Pose Estimator-subsection}. In addition, the proposed network can be easily extended to other dexterous hands and tactile sensors by simply changing the number of nodes in the graph representation.

\begin{table*}[t]
\renewcommand\arraystretch{1.5}
\caption{Experimental result on 6D pose estimation with positional error and angular error metric.}  
\begin{center}
\begin{tabular}{|l|c|c|c|c|c|c|c|c|} 
\hline
Metric                & \multicolumn{4}{c|}{Positional error (cm)}                                                                                                                                   & \multicolumn{4}{c|}{Angular error (rad)}                                                                                                                                      \\ 
\hline
Method                & \multicolumn{1}{l|}{Tactile} & \multicolumn{1}{l|}{Vision} & \multicolumn{1}{l|}{\begin{tabular}[c]{@{}l@{}}Visuo-tactile\end{tabular}} & \multicolumn{1}{l|}{SelectLSTM} & \multicolumn{1}{l|}{Tactile} & \multicolumn{1}{l|}{Vision} & \multicolumn{1}{l|}{\begin{tabular}[c]{@{}l@{}}Visuo-tactile\end{tabular}} & \multicolumn{1}{l|}{SelectLSTM}  \\ 
\hline
003\_cracker\_box     & 0.9                          & 0.6                         & 0.5                                                                           & \textbf{0.4}                    & 0.113                        & 0.078                       & 0.062                                                                         & \textbf{0.054}                   \\ 
\hline
004\_sugar\_box       & 1.0                          & 0.8                         & 0.6                                                                           & \textbf{0.5}                    & 0.383                        & 0.105                       & 0.098                                                                         & \textbf{0.09}                    \\ 
\hline
006\_mustard\_bottle  & 1.1                          & 1.5                         & 1.2                                                                           & \textbf{0.8}                    & 0.124                        & 0.112                       & 0.113                                                                         & \textbf{0.092}                   \\ 
\hline
016\_pear             & 1.5                          & 1.7                         & 0.9                                                                           & \textbf{0.7}                    & 0.334                        & 0.32                        & 0.135                                                                         & \textbf{0.132}                   \\ 
\hline
021\_bleach\_cleanser & 0.7                          & 0.7                         & 0.6                                                                           & \textbf{0.5}                    & 0.182                        & 0.116                       & 0.118                                                                         & \textbf{0.083}                   \\ 
\hline
033\_spatula          & 2.6                          & 2.1                         & 1.6                                                                           & \textbf{1.4}                    & 0.614                        & 0.34                        & 0.431                                                                         & \textbf{0.264}                   \\ 
\hline
065\_f-cups           & 0.9                          & 0.8                         & 0.6                                                                           & \textbf{0.5}                    & 0.228                        & 0.131                       & 0.122                                                                         & \textbf{0.099}                   \\
\hline
\end{tabular}
\end{center}
\label{table:1}
\end{table*}

\subsection{6D Pose Estimator}\label{6D Pose Estimator-subsection}

As mentioned above, we use the tactile-only, vision-only, and visuo-tactile fusion methods to estimate the candidate object 6D poses. The overall pipeline is shown in Fig. \ref{fig4}. Specifically, for tactile data, based on the graph representation in \ref{Graph-Representation-subsection}, we employ NNConv \cite{gilmer2017neural} to transfer 19-dim tactile data on the edges into nodes and use the GCNConv \cite{kipf2016semi} to extract the features from tactile data and expand the node features. Finally, we transform the features on the nodes into a single dimension. We also use the fully connected network to extract the features again and finally output a 7-dim vector, including 3-dim XYZ coordinates and 4-dim quaternion orientation to represent the 6D pose of the object. For vision data, we use a ResNet18 network as the backbone for feature extraction. In detail, a bounding box around the object is cropped by its mask, then the cropped RGB-D data is reshaped into 128*128*4 and fed into the ResNet18 network which outputs a 7-dim vector as object pose. Other networks for 6D pose estimation based on vision data are also possible, such as MoreFusion \cite{wada2020morefusion} and DenseFusion \cite{wang2019densefusion}, etc. For tactile and visual information fusion, we extract the merged feature vectors by concatenating the feature from the tactile and vision network. After the concatenation, we still use another fully connected network (1 layer) to output a 6D pose of the object. Regarding the loss function, we use $L_A$ loss functions similar to those in PoseCNN and DenseFusion for quaternion orientation regression \cite{xiang2018posecnn,wang2019densefusion}. 

\begin{equation}
\begin{aligned}
L_A =\frac{1}{2n}\sum_{x\in \mathcal{M}}||(R(\hat{q})x+\hat{t})-(R(q)x+t)||^2~
\end{aligned}
\label{eq1}
\end{equation}
where $\mathcal{M}$ stands for the set of 3D model points and n is the number of points. $R(\hat{q})$ and $R(q)$ denote the rotation matrices computed from the quaternion prediction and the ground truth rotation matrix, $\hat{t}$ and $t$ denote the predicted translation and the ground truth translation, respectively.

\subsection{SelectLSTM}
In real data, the result of the merging of multiple modalities is not necessarily better than either of the single input channels, especially when one channel is temporarily occluded or even corrupted. Therefore, we output the selection of results, ultimately improving the overall network’s performance. To realize that, we propose SelectLSTM. This recursive network performs sequence recognition on the estimated poses to avoid a miscalculation of the object pose caused by the sudden occlusion of the palm or the tactile sensors not touching the object. As illustrated in Fig. \ref{fig4}, the input to our SelectLSTM network is composed of a 21-dimensional vector obtained by concatenating the outputs of three pose estimators and a 256-dimensional multi-modal fusion feature vector, consisting of a 128-dimensional tactile vector and a 128-dimensional vision vector. The output of the SelectLSTM is a 3-dimensional vector that represents the confidence of three estimated poses, with the correct prediction being the pose whose average model distance (ADD) defined in Eq. (\ref{eq1}) is closest to the ground truth pose among the three predicted poses. The SelectLSTM comprises three layers of LSTM and a 256-dimensional feature vector serving as the hidden layer, and we train the network by sampling 20 consecutive time series.

\section{EXPERIMENTS and RESULTS}

\subsection{The Effect of SelectLSTM}\label{SelectLSTM-subsection}
In the experiment, we use the ObjectInHand dataset to evaluate the improvement of the SelectLSTM. We divide the dataset into 60\% for training and 40\% for testing. We perform tactile-only, vision-only, and visuo-tactile fusion, and SelectLSTM on the ObjectInHand dataset. To evaluate the performance of our proposed method, we adopt two metrics proposed in \cite{dikhale2022visuotactile}: position error and angular error. The position error is the Euclidean distance between estimated position $\hat{T}$ and ground truth position $T$. The angular error is calculated using the angle between two unit quaternions, as shown in equation (2). Here, $\theta$ represents the angular error, $\hat{q}$ is the estimated quaternion from the network, $q$ is the ground truth quaternion, and $\langle \hat{q}, q \rangle$ is the inner product of the two quaternions.

\begin{equation}
\begin{aligned}
\theta = \cos^{-1} (2\langle \hat{q}, q \rangle^2-1)
\end{aligned}
\label{eq2}
\end{equation}
The results are presented in Table \ref{table:1}. Through the subsequent selection method, the accuracy of SelectLSTM on the overall dataset is the best. In terms of positional error, SelectLSTM outperforms tactile-only, vision-only, and visuo-tactile merge methods by 44.8\%, 41.5\%, and 20.0\%, respectively. For angular error, SelectLSTM outperforms tactile-only, vision-only, and visuo-tactile merge methods by 58.8\%, 32.3\%, and 24.6\%, respectively. These findings confirm the advantages of our proposed approach in improving the accuracy of in-hand object pose estimation.

\subsection{The Effect on Occlusion Rates and Number of Contact Tactile Fingers}
\begin{figure}[!b]
\centerline{\includegraphics[width=3.5in]{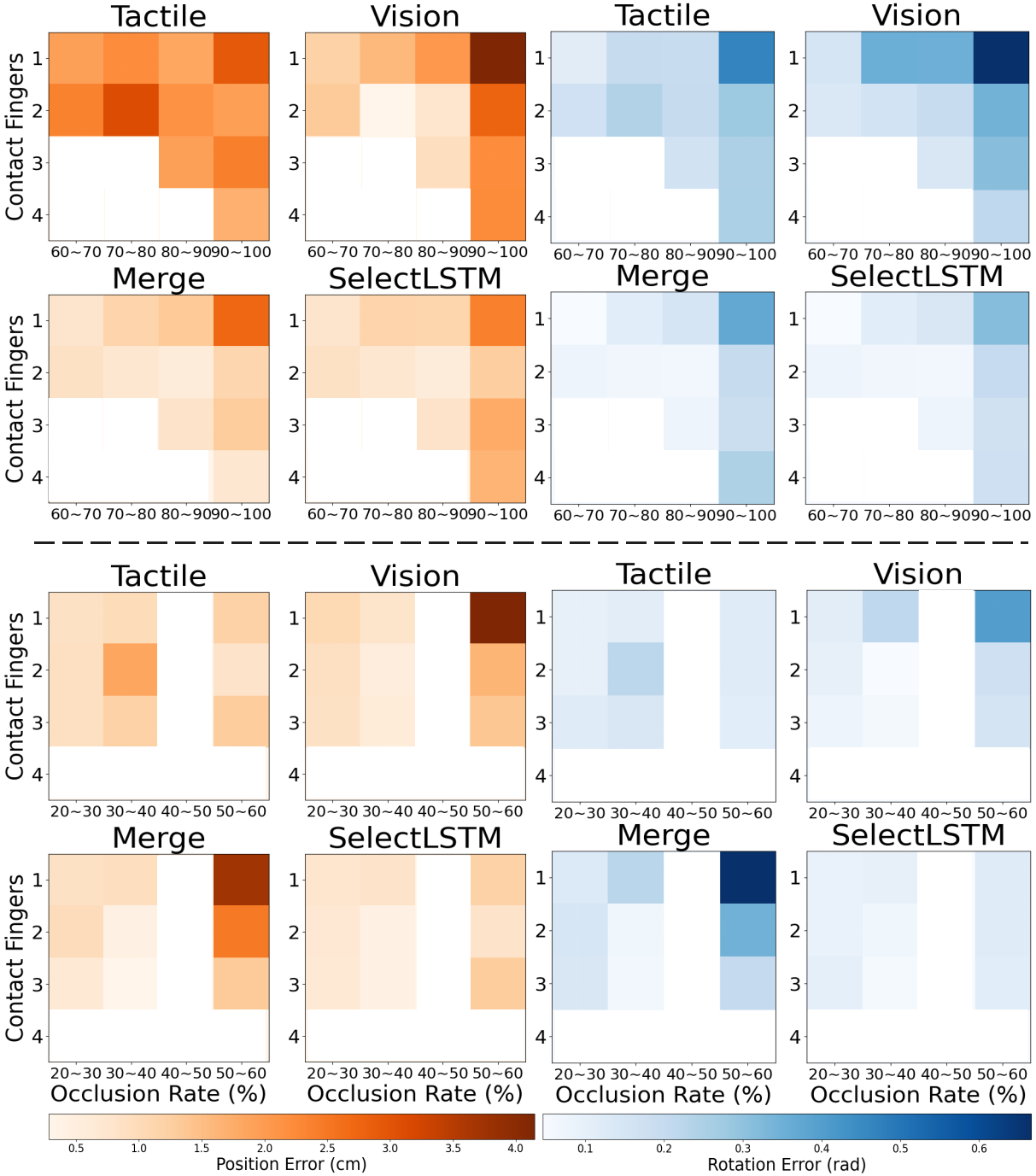}}
\caption{Angular error and positional error on the experiments with different methods on different occlusion rates and different number of contacting tactile fingers. The darker the block the higher the error; white blocks refer to no data available. The figures above the dashed line correspond to the pear, and figures below the dashed line correspond to the mustard bottle.}
\label{fig6}
\end{figure}

We evaluated the impact of occlusion rates and the number of contacting tactile fingers on the PoseFusion performance by selecting one small object (pear) and one large object (mustard bottle) from our dataset. For each object, we created a $4\times4$ matrix in Fig. \ref{fig6}, with four levels of occlusion rates and four numbers of tactile fingers contacting the object. The color of each cell indicates the corresponding error rate based on two metrics defined in \ref{SelectLSTM-subsection}, with the pear matrix above and the mustard bottle matrix below. The results show that as the occlusion rate increases and the number of contacting tactile fingers decreases, all object pose estimation methods experience a significant decrease, however, SelectLSTM consistently outperforms the other methods. Notably, for the mustard object, the visuo-tactile fusion method exhibits a decline in performance when the vision-only method is heavily degraded, while SelectLSTM selects the tactile-only method as the optimal result, avoiding the problem of modal collapse.

\subsection{Experimental Result on Occlusion and Tactile Missing}

\begin{figure}[!b]
\centerline{\includegraphics[width=3.5in]{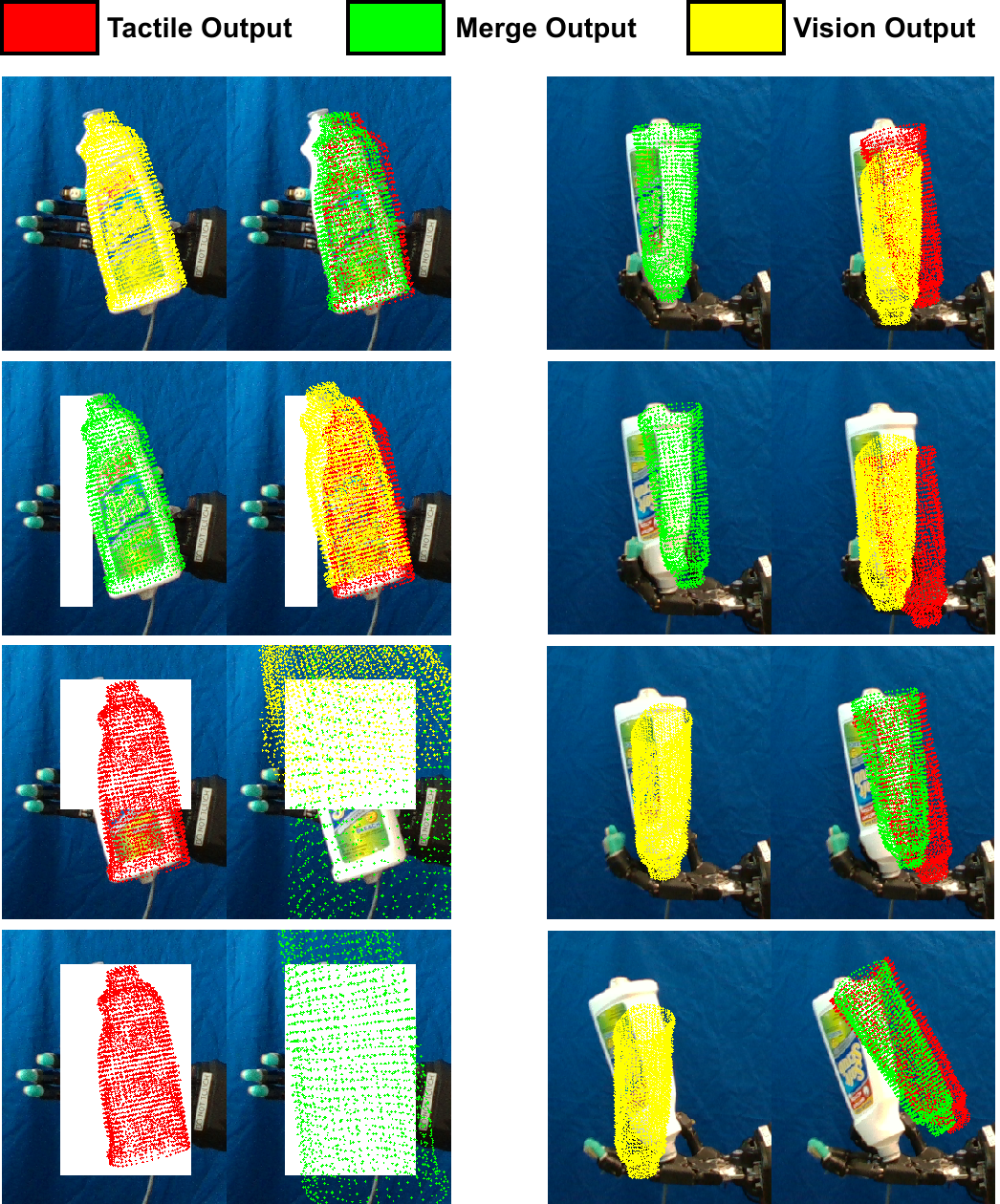}}
\caption{Pose estimation and selection under the corrupted data scenarios. The pose estimate results from tactile-only, vision-only and visuo-tactile fusion are depicted in red, yellow, and green, respectively. In the two-column data, the left shows the selected pose estimation result by SelectLSTM, while the right shows the pose estimation results based on the other three data types.}
\label{fig7}
\end{figure}

To further demonstrate the effectiveness of our SelectLSTM, we intentionally added noise to both vision and tactile data. For vision data, a white pixel block was manually added to gradually occlude the object. For tactile data, the object was placed on the palm of the robotic hand without contact with the tactile sensor. Subsequently, we fed the noise-added data into the trained model to observe its corresponding outputs, the specific results are shown in Fig. \ref{fig7}. The first column shows that the estimated pose by the vision-only and visuo-tactile fusion methods gradually drifts as the occlusion area grows. However, the object pose, which was selected by the SelectLSTM, estimated by tactile-only data remains stable. In the second column, the estimated object pose by tactile-only or visuo-tactile fusion exhibits motion even though there is no direct contact between the fingers, while the estimation result of pure vision can be well maintained. These results demonstrate the robustness and effectiveness of our SelectLSTM approach in handling noisy data and selecting reliable 6D object pose estimates.

\section{Discussion and CONCLUSION}

\subsection{Discussion}
Despite the experimental results verifying our hypothesis, there are still some limitations to our work: regarding datasets, we currently use the magnetic sensor-based method for ground truth pose acquisition, but methods used in other works \cite{hampali2020honnotate} for pose acquisition may avoid the influence of the sensor wire. Secondly, since we focus on the post-fusion of the network outputs, our current visuo-tactile fusion method just simply connects feature vectors, some advanced fusion methods such as Gate or Transformer models can be applied to the network further to improve the performance \cite{anzai2020deep} \cite{bai2022transfusion}.

\subsection{Conclusion}
In this study, we have introduced a new dataset for 6D object pose estimation in the context of dexterous hand manipulation using visual-based teleoperation. To address the challenge of single-modal corruption that often occurs in real-world scenarios, we have proposed SelectLSTM, which can make a selection at the output layer to improve the pose estimate results. Our dataset is publicly available on our project website, and our experiments have demonstrated that SelectLSTM can effectively decrease the pose estimation error. In future work, we aim to generate realistic data in a simulator to allow the creation of much larger training sets
to further improve the performance of our deep networks.

\bibliographystyle{IEEEtran}

\bibliography{reference}

\end{document}